\documentclass[twocolumn]{article}
\pdfoutput=1 % pdflatex must be used
\usepackage{amsmath,graphicx}
\PassOptionsToPackage{dvipdfmx}{graphicx} %or dvipdfm depending on the tex system

\usepackage{subcaption}

\usepackage[affil-it]{authblk} % something with the authors and affiliations.

\usepackage{booktabs}
\usepackage{float}

\usepackage[section]{placeins} % dont float outside section

\usepackage{pgf}

\usepackage{hyperref}
\hypersetup{
    colorlinks=true,
    linkcolor=blue,
    filecolor=magenta,      
    urlcolor=cyan,
}
\usepackage{times}

\renewcommand{\textcolor}[2]{}
\renewcommand{\colorbox}[2]{}

\title{RootPainter3D: Interactive-machine-learning enables rapid and accurate contouring for radiotherapy}

\author[1, 2, *]{Abraham George Smith}
\author[1, 2]{Jens Petersen}
\author[3]{Cynthia Terrones-Campos}
\author[2,4]{Anne Kiil Berthelsen}
\author[1, 2]{Nora Jarrett Forbes}
\author[1]{Sune Darkner}
\author[2]{Lena Specht}
\author[2]{Ivan Richter Vogelius}

\affil[1]{\small{Department of Computer Science, University of Copenhagen}}

\affil[2]{\small{Department of Oncology, Rigshospitalet, University of Copenhagen}}

\affil[3]{\small{Department of Infectious Diseases, Rigshospitalet, University of Copenhagen}}

\affil[4]{\small{Department of Clinical Physiology, Rigshospitalet,
                University of Copenhagen}}

\affil[*]{ags@di.ku.dk}

\date{}% leave blank to omit date.

\begin{document}
%\ninept
\maketitle
\begin{abstract}

%“For Research Articles, Technical Notes, and Medical Physics Letters, a structured abstract is required that consists of 4 parts: Purpose, Methods, Results and Conclusions. For Research Articles, that abstract should not exceed 500 words. A limit of 300 words applies to Technical Notes and Medical Physics Letters."

Organ-at-risk contouring is still a bottleneck in radiotherapy, with many deep
learning methods falling short of promised results when evaluated on clinical
data. We investigate the accuracy and time-savings resulting from the use of an
interactive-machine-learning  method for an organ-at-risk contouring task. We
compare the method to the Eclipse contouring software and find strong agreement
with manual delineations, with a dice score of 0.95. The annotations created
using corrective-annotation also take less time to create as more images are
annotated, resulting in substantial time savings compared to manual methods,
with hearts that take 2 minutes and 2 seconds to delineate on average, after
923 images have been delineated, compared to 7 minutes and 1 seconds when
delineating manually.  Our experiment demonstrates that
interactive-machine-learning with corrective-annotation provides a fast and
accessible way for non computer-scientists to train deep-learning models to
segment their own structures of interest as part of routine clinical workflows.

Source code is available
at \href{https://github.com/Abe404/RootPainter3D}{this HTTPS URL}.

\end{abstract}

%\begin{keywords}
%Deep-learning, Segmentation 
%\end{keywords}
%
\section*{Introduction}
\label{sec:intro}

Half of all cancer patients receive radiotherapy \cite{delaney_role_2005},
which is associated with a range of dose dependent side effects
\cite{barazzuol_prevention_2020}. Effective mitigation of these side effects
requires accurate delineation of organs-at-risk, such as the heart and
oesophagus \cite{ezzell_guidance_2003, mackie_image_2003}. Manual delineation is
still widely used but time-consuming in comparison to automated methods
\cite{tang_clinically_2019} and subject to large inter-observer variation
\cite{joskowicz_inter-observer_2019, yang_statistical_2012-1}. A review of
auto-segmentation methods for radiotherapy is presented by
\cite{cardenas_advances_2019}, indicating deep-learning methods and
convolutional neural networks (CNN) in particular as representing the
state-of-the-art. A survey of deep-learning for radiotherapy is presented by
\cite{meyer_survey_2018}, including explanations of machine learning, deep
learning, and CNNs.

\begin{figure}
    \centering
\includegraphics[width=0.5\textwidth]{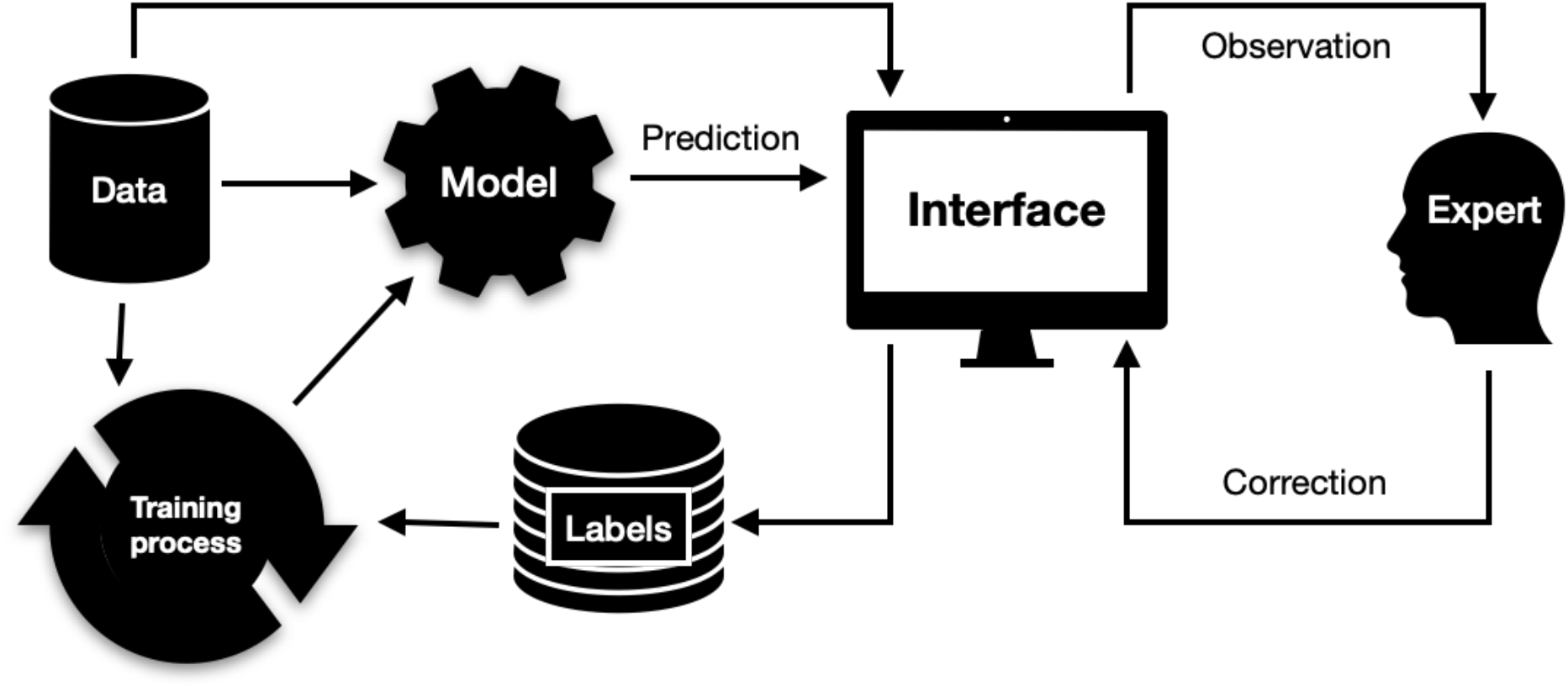}
  \caption{Interactive machine learning with corrective annotation puts the human-in-the-loop
           during the model training process.}
\vspace{-0.5cm}
\label{fig:iml_diagram}
\end{figure}

Although CNNs exhibit impressive performance when the training and testing data
are drawn from the same distribution \cite{milletari_v-net:_2016,
ibragimov_segmentation_2017, tang_clinically_2019, feng_deep_2019,
um_multiple_2020}, variations specific to on-site clinical data may result in
decreased performance \cite{gibson_inter-site_2018, ghafoorian_transfer_2017,
schreier_generalization_2020}.  For example \cite{feng_improving_2020} found
that organ deformations due to an abdominal compression technique impaired
the performance of an externally trained CNN model.

Training models on-site is a potential solution but can be challenging as
training neural networks involves time-consuming trial and error
\cite{meyer_survey_2018} and hiring the appropriate experts is associated with
high costs. A lack of large and high-quality publicly available datasets
compounds the problem \cite{vogelius_harnessing_2020} as annotating large
enough datasets for training deep-learning models with purely on-site data may
be infeasible.

Corrective-annotation is an annotation sparsification strategy that results in
a sub-region of each image being labelled. \cite{zheng_annotation_2020}
demonstrate, using a 3D heart segmentation dataset, that
annotation-sparsification allows deep-learning models to be trained using just
20\% of the labelled data whilst obtaining similar performance. As opposed to
\cite{zheng_annotation_2020} who rely on automatic methods to infer which
regions of an image are most useful for labelling, corrective-annotation
utilises human feedback (Figure \ref{fig:iml_diagram}) to identify problematic
regions of the image for the model \cite{gonda_icon_2017,
smith_rootpainter_2020, ho_deep_2020-3, kontogianni_continuous_2020}.  Methods
such as corrective-annotation which utilise a human-in-the-loop to improve
machine learning implementations are known as interactive machine learning
(IML) \cite{holzinger_interactive_2016}.

% Prior work discussion - keep in intro.
In prior work, corrective-annotation IML methods have been used to train 
 deep-learning segmentation models for a diverse
array of modalities, including 2D plant photographs
\cite{smith_rootpainter_2020, han_digging_2020, denison_legume-imposed_2021},
laser ablation tomography \cite{elias_interactive_2021}, histopathology
\cite{ho_deep_2020-4} and 3D X-ray CT images of methane in sand
\cite{alvarez-borges_u-net_2021}.

To the best of our knowledge, no prior work has evaluated an IML approach
applied to medical X-ray CT data.
Therefore we implement an IML system for X-ray CT data and investigate its
effectiveness for organ-at-risk segmentation.

% Objective / Hypothesis
We hypothesise that (1) semi-automatic contouring via an IML
corrective-annotation method will provide similar accuracy to existing manual
methods; And (2) that it will offer continuous improvements in contouring time
as more images are annotated, eventually leading to significant time savings.
       
\section*{Method}
\label{sec:method}

We use a corrective-annotation approach to contour a large dataset of hearts by
having a physician correct all generated contours during training (Figure
\ref{fig:iml_diagram}). The process is semi-automatic, with the assisting model
continuously learning as more images are annotated.

\subsection*{Dataset}
\label{sec:dataset}

We used X-ray CT scans from a cohort of patients that had been collated for a
study on the association between mean heart dose and radiation-induced
lymphopenia \cite{terrones-campos_prediction_nodate}. The CT scans were from
patients who had started to receive radiotherapy to the chest region between
2009 and 2016 at Rigshospitalet in Denmark. The cohort was restricted to
patients who were at least 18 years old, had dosimetric data available, a solid
malignant tumour (excluding lymphoma) and a blood count both before and after
receiving radiotherapy. This resulted in 933 X-ray CT scans from 923 patients,
of which 308 had breast cancer, 291 had esophageal cancer, 56 had
small-cell-lung-carcinoma and 268 had non-small-cell lung carcinoma.  The 933
images had varying slice thicknesses. 209 images had a slice thickness of 3mm,
721 had a slice thickess of 2mm , and 3 had a slice thickness of 1mm. The
number of voxels in each dimension varied.  The total number of slices (depth)
varied from 96 to 489 with a mean of 189. The width and height of the axial
plane ranged from 512 to 658, with a mean of 537 voxels. No pre-processing was
performed to normalize the images in any way. All files were saved in 
the gzip compressed NIfTI file format with extentsion .nii.gz.
% This citaiton is just too long.. so many authors. Maybe try again w
% to facilitate
% convenient loading with nibabel library \cite{brett_nipynibabel_2020}.

\subsection*{Software Implementation}

We implemented RootPainter3D, which is based on RootPainter, an open-source
corrective-annotation software application that utilises a client-server
architecture and makes the necessary operations to train and use a
deep-learning model for image segmentation available via a cross-platform
graphical user interface \cite{smith_rootpainter_2020}.

The original RootPainter uses a variant of U-Net \cite{navab_u-net_2015}
modified to utilise group norm \cite{wu_group_2018} and residual connections
\cite{he_deep_2015}. Compared to the 2D version we modify the software in
several ways. The 2D convolutional layers were converted to 3D, resulting in a
variant of the 3D U-Net architecture which is known to perform well for
organ-at-risk segmentation \cite{feng_deep_2019} and sparse data
\cite{cicek_3d_2016-1}. The interface was modified to allow contrast settings
to be adjusted, navigation and annotation of 3D images, with viewing enabled in
both sagittal and axial views simultaneously. The data augmentation was
removed. Although it is claimed data augmentation is critical to achieving
favourable generalisation performance \cite{schwobel_probabilistic_2020},
results indicate the advantages of data-augmentation can be inconsistent and
dataset specific \cite{isensee_nnu-net_nodate}.

RootPainter3D allows a user to inspect model predictions on a dataset that they
work through sequentially - one image at a time. For each image, they initially
define a bounding box in order to obtain predictions for a region containing
the organ of interest. The user is able to assign corrections to the model
predictions given for the defined region. When the user clicks 'save and next'
in the interface the current annotation is saved to disk in a folder of
annotations which is shared with a remote server with a more powerful graphics
processing unit (GPU). The server component of the software continuously trains
a 3D U-Net on the available annotations using stochastic gradient descent. For
RootPainter3D, batches of 4 annotations, with their associated image regions
are sampled from the annotation folder without replacement. The regions of the
images used for training are only those where user-annotations are assigned.
Supervised training is used but the annotations are sparse. Only the defined
regions (what the user has specified as either foreground or background) are
used for computing the loss which is used for updating the model weights. 
\begin{figure}
\centering
\begin{subfigure}{0.47\textwidth}
  \centering
  \includegraphics[width=\textwidth]{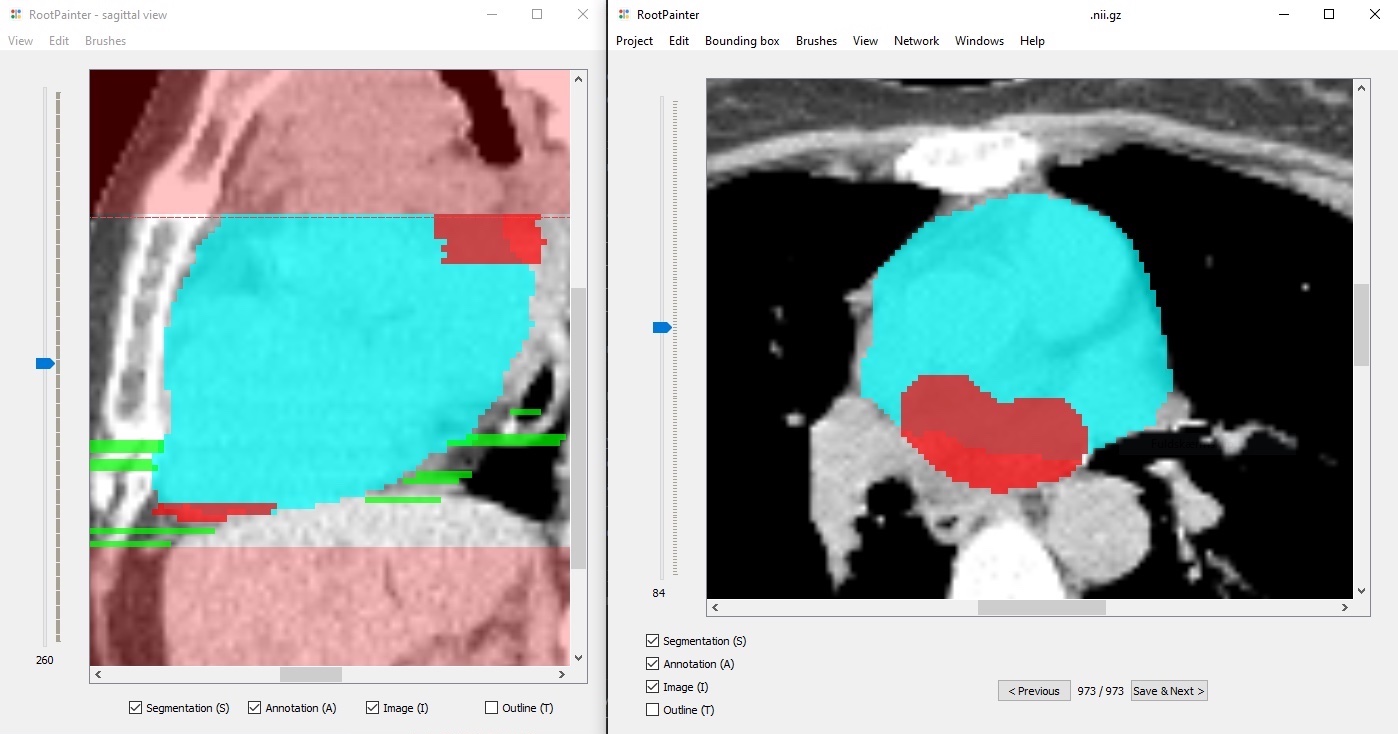}
    \caption{Screenshot of RootPainter3D software showing segmentation in blue
and annotation in red for corrections to false negative regions and green for
corrections to false positive regions. The area in light red is outside of the
bounding box and is not part of the predicted or corrected region. The dashed
red line shows the position of the axial slice in the sagittal view. This dashed line 
was added after the experiment in this study was completed based on user-feedback.} \label{fig:screenshot_01}
\end{subfigure}
\begin{subfigure}{0.47\textwidth}
  \centering
  \includegraphics[width=\textwidth]{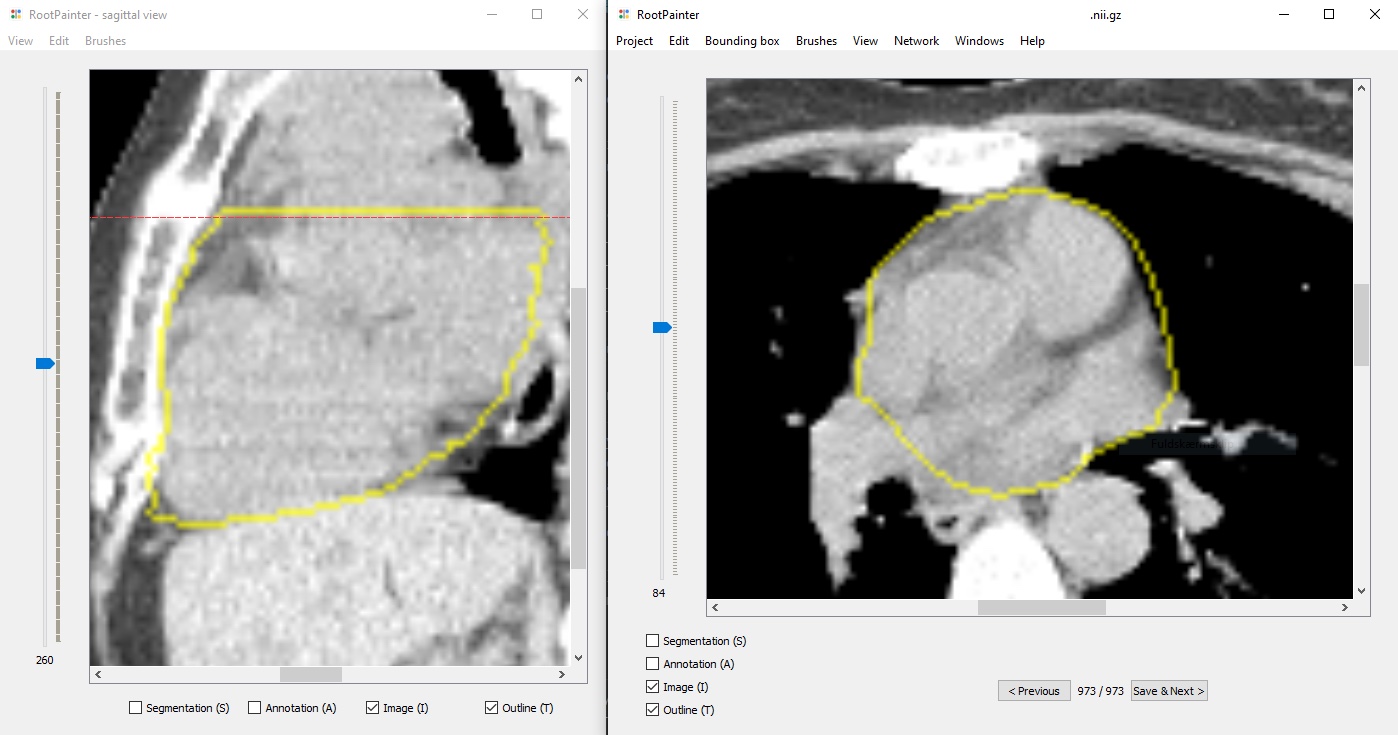}
  \caption{Screenshot of RootPainter3D software showing the outline view with segmentation and annotation hidden.}
  \label{fig:screenshot_02}
\end{subfigure}
    \caption{Screenshots of the RootPainter3D client software. Showing different views available for
    heart contouring. The axial view is shown on the right and sagittal view on the
    left. The user is able to change what data is shown via keyboard shortcuts or
    with the checkboxes shown in each viewer.}
\label{fig:screenshots}
\end{figure}

The contrast settings (see \nameref{sec:contouring} for more details).
can be changed to a preset option in the view menu, such
as mediastinal, which was used for our experiments. 

% IRV: Too much detail (consider putting in the readme or something)
%The presets can be added,
%edited and removed by editing the \textit{root\_painter\_settings.json} file
%which is automatically created in a users home folder when the RootPainter3D
%client program is first ran. 

\subsection*{Contouring procedure}
\label{sec:contouring}

To enable comparison to a manual method widely used in the clinic, 20 hearts
were also delineated using the Eclipse treatment planning software from Varian
Medical Systems, Inc.

From the 933 heart CT scans, the annotator delineated the first 10 in Eclipse
and then all 933 in RootPainter3D and then the last 10 in Eclipse. This was
done to allow the dice score to be computed between the contours done in
RootPainter3D and Eclipse, and also to compare delineation time between
RootPainter3D and Eclipse at the start and end of the RootPainter3D model
training process.

Contouring in both Eclipse and RootPainter3D was done using the mediastinal
hounsfield unit (HU) window which ranges from -125 HU to 250 HU. For both
software applications, annotation can be performed in both the sagittal and
axial views but for our experiment all annotation was assigned in the axial
view. In RootPainter3D the user is able to view the model's initial
segmentation in blue (Figure \ref{fig:screenshot_01}). They can annotate
foreground (heart) regions in red and background (not heart) regions in green
(Figure \ref{fig:screenshot_01}). A 3D segmentation consists of a model
prediction for each voxel in an image. If annotating correctively, the red
foreground annotation should correspond to model predictions that were false
negatives, and the green background regions should be targeted towards false
positive model predictions. Taking corrections into account, the corrected
contour can be viewed using the outline view (Figure \ref{fig:screenshot_02}).

When contouring in Eclipse, every third slice was contoured and then
interpolation was used to join the slices, as this is a standard clinical
procedure used to save time compared to contouring all slices. The same trained
physician delineated all hearts in both RootPainter3D and Eclipse using the
Danish guidelines for whole heart contouring \cite{milo_delineation_2020}.

In order to ensure the physician's familiarity with both software applications
prior to the experiment, 20 hearts that were not used as part of the experiment
were contoured in both applications. 

\subsection*{Training}

The training process for neural networks is a procedure where the network
weights and biases are iteratively updated by a gradient descent algorithm,
which compares the network outputs with the desired output
\cite{lecun_deep_2015}.

When a user completes a delineation in the RootPainter3D client software, the
annotation is saved to disk and training is started automatically by sending an
instruction to the server.

% We decided this was unnesessary
%\textit{start\_training} instruction is saved
%to the instructions directory in the \textit{root\_painter\_sync} directory.
%This is synced with the server using an existing network file system.

%On the server, the RootPainter3D trainer application watches the instructions
%directory for new files. Once it receives an instruction such as
%\textit{start\_training}, it executes the associated python function.

The training routine, once started, first creates a model using Kaiming
initialisation to assign random parameters \cite{he_delving_2015}, and then
trains that model by continuously iterating over all saved annotations.

The training procedure treats newly saved images similarly to to previously
saved images as all images are equally likely to be used in the next batch of
the training procedure. It loads the annotations from disk and uses these,
combined with the image data, to train a model to predict the annotated regions
correctly.  For the RootPainter3D trainer, regions that are not annotated are
set to 0 in both the network predictions and annotation. This means only the
regions of the image explicitly annotated as either foreground or background by
the annotator are used to compute the loss.
% a measure of the distance between the network predictions and user-defined
% annotation. 
The loss is then used to update the network weights. The loss function used in
RootPainter3D is a combination of cross-entropy and dice loss, taken from
\cite{smith_segmentation_2020}.

A GPU is required for achieving optimal performance when training large
parameter models such as deep neural networks.  As the entire dataset cannot
fit in the memory of the GPU, stochastic gradient descent is used to make
network updates based on a subset (known as a batch) of the full available
dataset. In our case the network is trained with a batch size of 4. We used two
NVIDIA Titan RTX GPUs, with 24GB of memory each, using a
data-parallel \cite{dean_large_nodate} approach.  Due to GPU memory
constraints, instead of using the full images, a sub-region (patch) is sampled
from a random location that contains annotation, from each of the randomly
sampled images in the training batch. The patch dimensions are 228, 228 and 52
voxels, corresponding to width, height and depth respectively.

For the RootPainter3D trainer application, once training has started, the
process is continuous, even when the client application is closed.  The trainer
application keeps track of a counter named \textit{epochs\_without\_progress}.
This counter is used to measure the length of time that the network has been
trained without obtaining a new high score on the validation data.

\subsection*{Validation}
\label{sec:validation}

In a machine learning context, the purpose of validation is to estimate model
performance on unseen data.  Networks with large capacity can overfit arbitrary
datasets \cite{zhang_understanding_2017}. Fortunately, they have a tendency to
learn smooth functions and overfitting can be conveniently mitigated with early
stopping \cite{caruana_overfitting_nodate}. Early stopping involves checking
model performance on a portion of the data not used for training. This allows
generalisation performance to be estimated and a snapshot of the model weights
to be taken before overfitting has substantially degraded performance.

Similarly to 2D RootPainter \cite{smith_rootpainter_2020}, a portion of the
annotated images are assigned to the validation set instead of the training
set. The first image is assigned to the training dataset, then the second image
is assigned to the validation dataset. From then on images are only assigned to
the validation dataset if the training dataset is at least 5 times as large as
the validation dataset.

The validation dataset is used for model selection, in a way equivalent to
early stopping.  An epoch typically refers to a full iteration over all
available training data. In this context we define epoch length in terms of the
number of examples sampled from the training data and have this automatically
adjusted in relation to the validation set size, which increases as more images
are annotated. If no validation images have been saved then the epoch length is
128. Otherwise the length is $max(64, 2 v)$ where $v$ is the number of patches
containing annotation in the validation set. Setting the length of the training
period based on the validation set size ensures validation time does not
overwhelm the training procedure, despite continuous addition of new cases.

At the end of each epoch the predictions for the model being trained are
computed on the validation set and a dice score is computed using the available
validation set annotations. If the model-in-training's dice score is higher
than the most recently saved model so far then it will be saved to disk and
used for generating the segmentation presented to the user in the client
user-interface.

\textit{epochs\_without\_progress} will get set to 0 in two cases. Firstly,
when new data is added to the training or validation datasets. Secondly, when a
new model is found which obtains a new high score on the validation set.

Network training procedures are stochastic and dependent on initial weights
\cite{atiya_how_1997}. 
% IRV suggested to remove the following. And I agree.
% The value of the network weights are unlikely to change
% a lot from their initial values \cite{li_learning_2019}, resulting in trained
% models that are restricted by the initial configuration
% \cite{jesus_effect_2020}.
Therefore as the first initial weights are unlikely
to be optimal, repeating the training procedure with different initial weights
is a simple way to obtain better results. For this reason we introduced a
restart procedure. The restart procedure was not functioning for the first 478
hearts, during which the network would stop after 60 epochs without progress.

The restarting behaviour, used for the latter portion of the experiment to
further boost network accuracy, would start training the network from scratch
after 60 \textit{epochs\_without\_progress} were reached. If the newly trained
model beat the best model on the validation set so far then it would be saved
(see \nameref{sec:validation} for more details).  That means if the system is
left unattended for long enough, it will likely find a new best model, and keep
trying relentlessly until it does.

\subsection*{Dice evaluation}

In order to evaluate the accuracy of the RootPainter3D contours before and
after user correction, we compared the dice of both the initial predicted
contour and the version after corrections with contours created manually in
Eclipse for the same 20 images.

For each model generated heart contour, we also compute the dice score between
the predicted heart contour and corrected version for 933 images. This approach
is similar to previous studies that use manually corrected model predictions as
ground-truth contours for evaluation \cite{feng_deep_2019,
gros_automatic_2019}.

\subsection*{Interaction logging and annotation duration}

To evaluate the extent to which the trained model assisted in reducing
annotation time, we asked the annotator to log the specific time periods when
they started and stopped annotating. We also automatically logged interaction
events with the user-interface including when the user saved an image, opened a
new image, moved to a different slice in the axial or sagittal views, changed
zoom settings and mouse down and mouse release events.

To compute the amount of time the annotator spent on each image, we filter to
interaction events inside their manually logged annotation period. For each
file, we start accumulating interaction time from when the user opens that file
until they open the next file. We also exclude periods of inactivity from our
duration computation, which we define as no interaction events for at least 20
seconds.

% ############ Maybe add this to the next paper as it didn't happen here.  For the esophagus, the
% the visiable axial slice in the sgiattal view, therefore we altered the
% software to display a red line showing the axial slice view over the sagittal
% image. 
% 
% For the esophagus, the physician also had problems delineating with the
% circle brush tool they often drew a 1 pixel outline and then slowly filled it
% in with ever increasing brush size as using a large brush would cause errors.
% To make this process faster we implemented a fill tool which will fill a
% closed region of the annotation with the current brush colour. The fill is
% enabled by pressing Alt when painting.

\subsection*{Impact on radiation dose}

We also compare the computed mean heart dose (Gy)
when using the initial segmentation compared to the corrected contour.

% IRV suggests to delete and I agree.
%These
%differences in dose absorption measurements provide indications of the utility
%of the assigned corrections and of the consequences of using the network for
%fully-automatic contouring for either dose-response modelling or in the clinic.

To compare dose absorption we first obtain a dose matrix for each scan from the
clinical dose plan. The dose matrix provides a cumulative level of radiation
dose absorbed by each voxel in the planning CT scan. We compute the mean heart
dose using the dose matrix combined with both the initial segmentation and
corrected structure independently. We then take the absolute difference between
the two mean heart dose measurements to allow us to plot dose deviation over
time as more hearts are annotated.

%Mean heart dose is of interest as it is frequently used in studies modelling
%dose-response relationships related cardiac toxicity
%\cite{bates_volumetric_2017, maraldo_cardiovascular_2015,
%van_nimwegen_risk_2017}.

\subsection*{Results}
\label{sec:results}

\subsubsection*{Duration}

\begin{figure*} \centering
\includegraphics[width=0.8\textwidth]{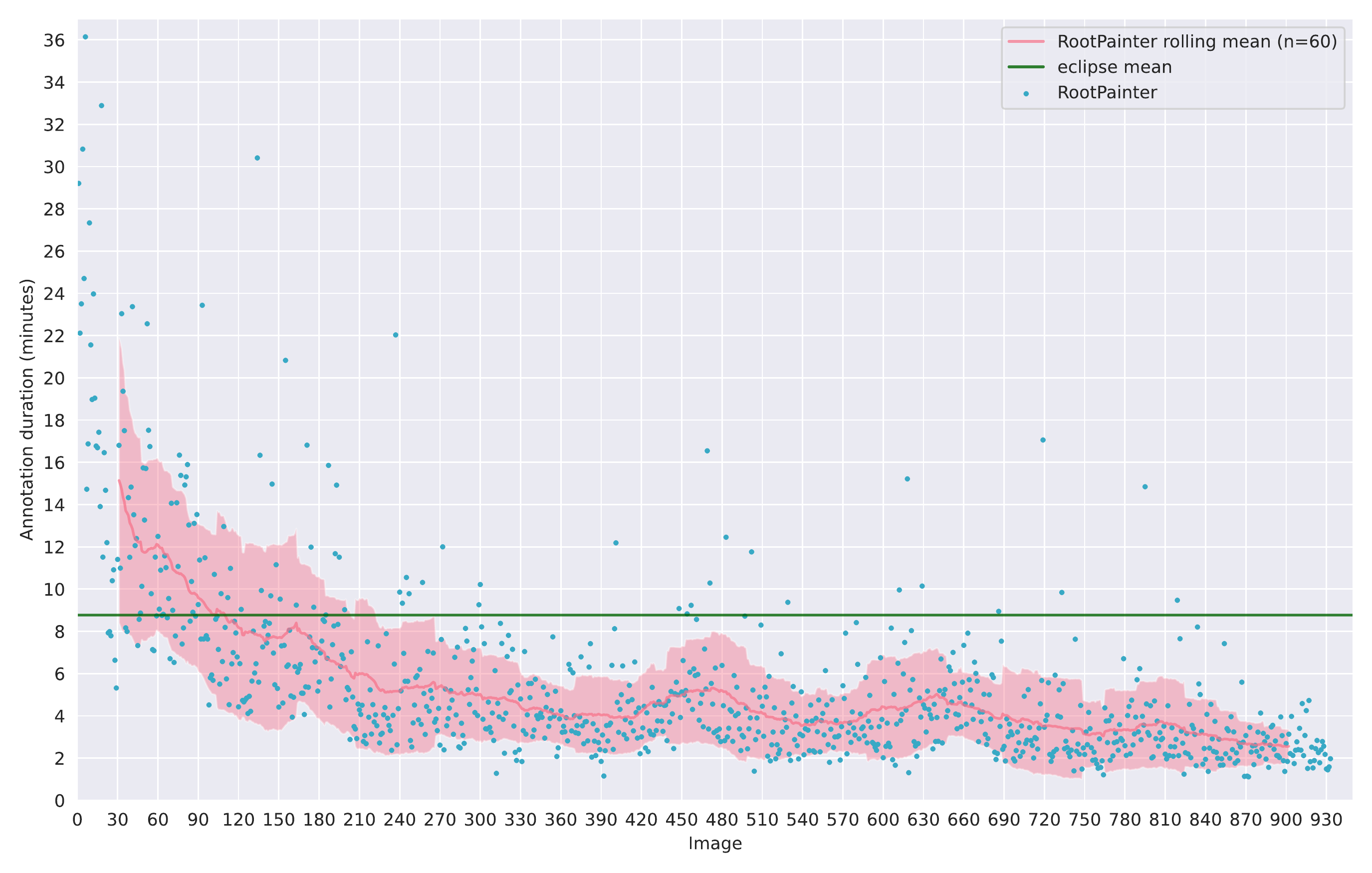}
\caption{Delineation duration for each of the 933 images shown in order of
delineation with the mean delineation duration of the 20 images contoured in
Eclipse. RootPainter3D per-image delineation duration reduces over time.}
\label{fig:duration_with_eclipse}
\end{figure*}

\begin{figure}
    \centering
\includegraphics[width=0.47\textwidth]{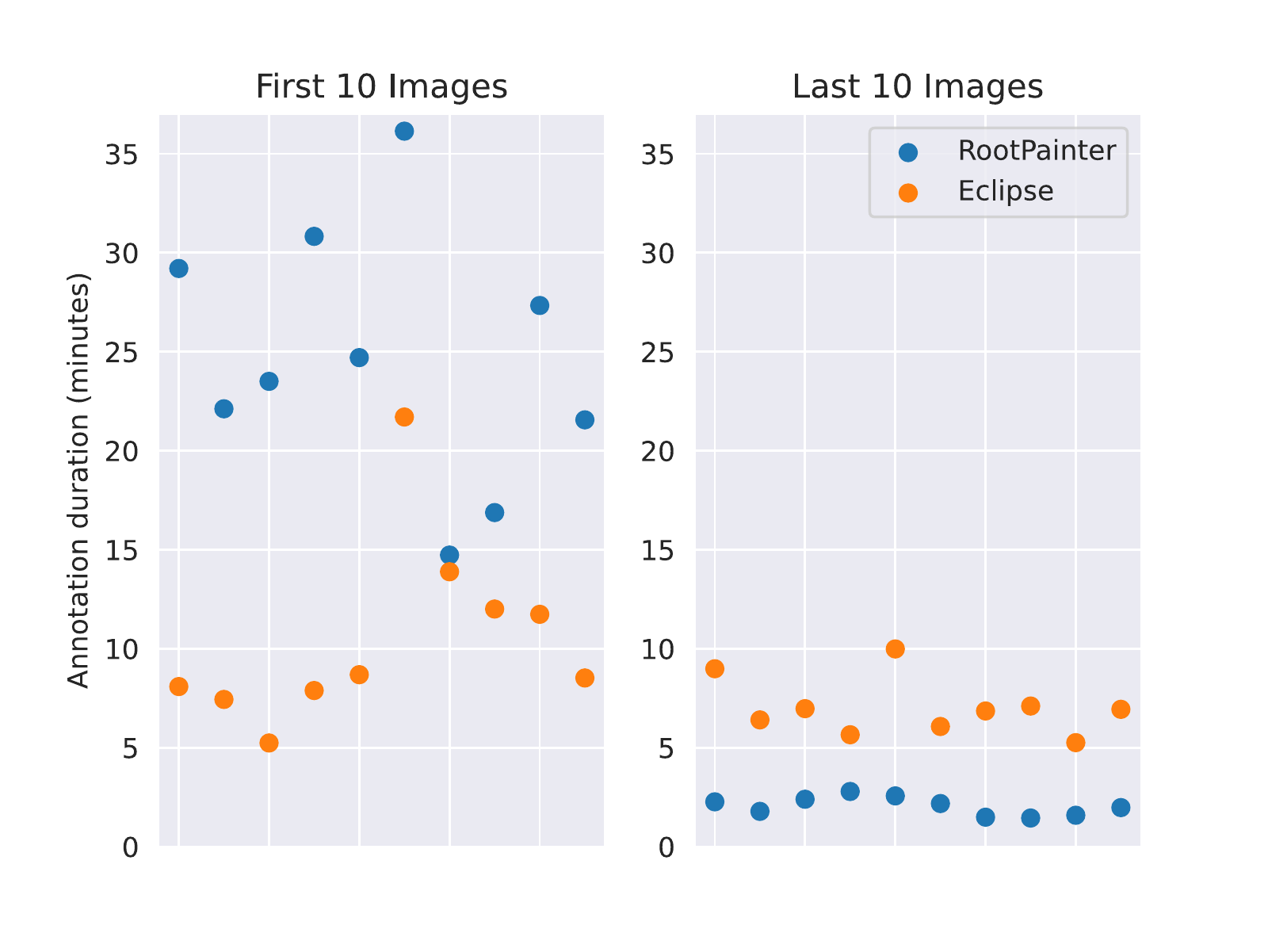}
  \caption{RootPainter3D and Eclipse delineation duration for the first 10 and
    last 10 images out of the 933. RootPainter3D is initially slower than Eclipse
    but then becomes significantly faster.  The last 10 hearts take  an average of
    2 minutes and 2 seconds to delineate in RootPainter3D compared to 7 minutes and
    1 seconds when using Eclipse. Delineation in RootPainter3D is over
    three times faster than delineation in Eclipse.
    }
\label{fig:two_plots_start_end}
\end{figure}

In order to evaluate the annotation speed of RootPainter3D, we report
delineation duration as a function of the number of images annotated (Figure
\ref{fig:duration_with_eclipse}). The first 10 and last 10 hearts are shown
separately to highlight both the differences with Eclipse and RootPainter3D
both before and after a period of extensive interactive annotation (Figure
\ref{fig:two_plots_start_end}). RootPainter3D is initially slower and then
becomes significantly faster after it has learned from more user corrections
(Figure \ref{fig:two_plots_start_end}).

After around 110 images, delineation duration in RootPainter3D is less than the
mean delineation of the 20 hearts contoured in Eclipse (Figure
\ref{fig:duration_with_eclipse}). With the exception of temporary fluctuations,
the RootPainter3D per-heart delineation duration continues to decrease as more
hearts are contoured (Figure \ref{fig:duration_with_eclipse}) becoming
substantially faster than the comparison manual method with the last 10 hearts
taking  an average of 2 minutes and 2 seconds to delineate compared to 7
minutes and 1 seconds when using Eclipse (Figure
\ref{fig:two_plots_start_end}). The RootPainter3D delineations also maintain
strong agreement with manual delineations of the same images (Table
\ref{table:last10dice}).

% Not required as data is shown in the plot
%\begin{table}[H]
%\caption{Delineation time in minutes for the last 10 hearts for both
%         RootPainter3D and Eclipse, showing that delineation in RootPainter3D is over
%         three times faster than Eclipse.}
%\centering
%\begin{tabular}{|l|l|l|}
%\hline
%     & Eclipse & RootPainter3D \\ \hline
%mean & 7.02    & 2.04          \\ \hline
%std  & 1.37    & 0.44         \\ \hline
%\end{tabular}
%\label{table:last10time}
%\end{table}

\subsubsection*{Accuracy}

\begin{figure*}
    \centering
\includegraphics[width=0.75\textwidth]{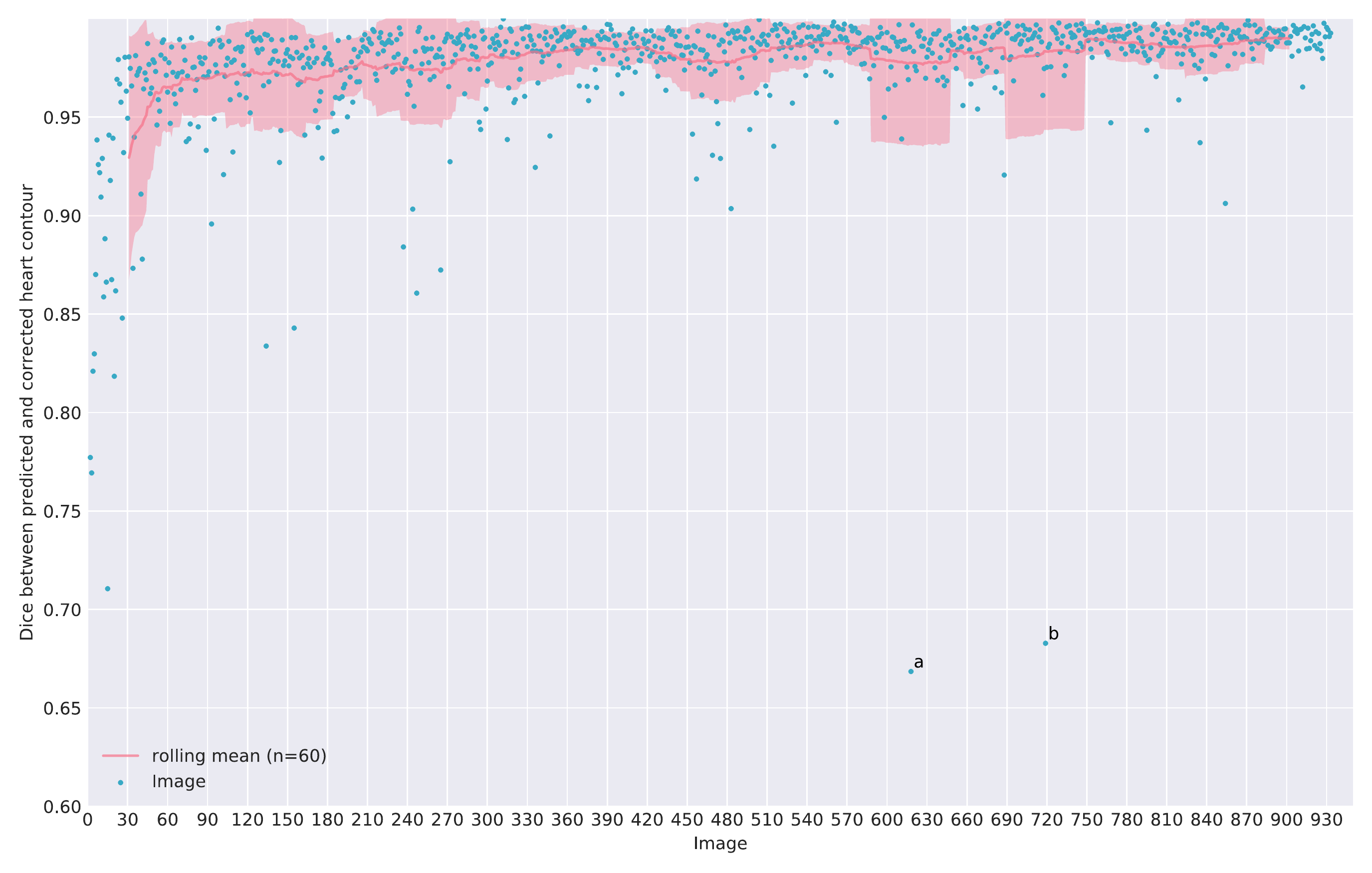}
\caption{The dice score for 933 images shown in delineation order. After the initial training period, the vast majority of dice scores are above 0.9, with only a few outliers dropping below. Outliers are labelled a and b.}
\label{fig:all_dice_06}
\end{figure*}

\begin{figure*}
  \centering
\includegraphics[width=0.75\textwidth]{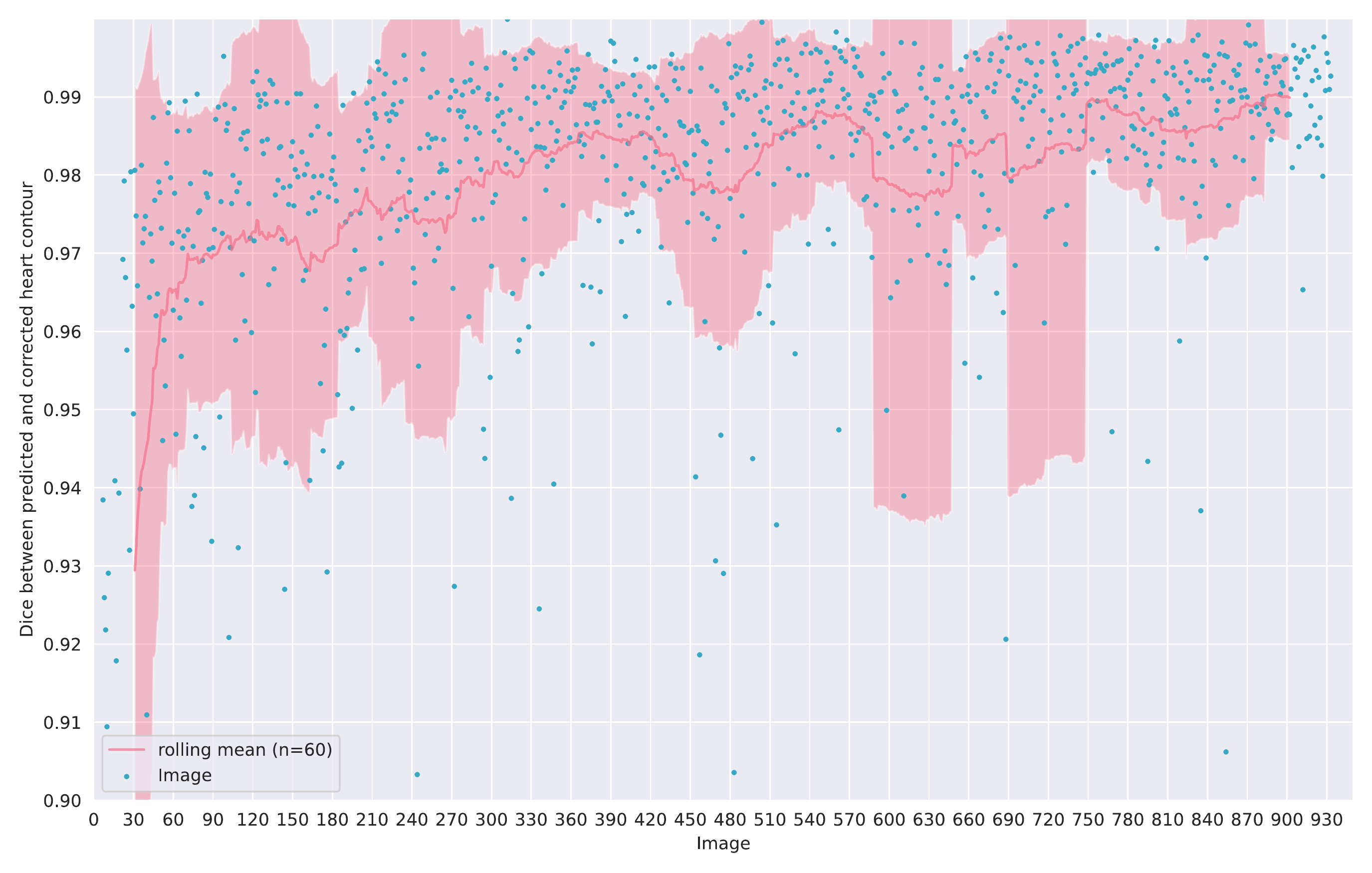}
  \caption{Dice score computed using the initial predicted image and the
corrected version. Y-axis restricted to 0.9 to 1.0 to show improvement in mean
over time. The mean dice score has fluctuations, but shows a general trend of improvement
throughout the experiment.}
\label{fig:all_dice_09}
\end{figure*}

The dice score between the initial predicted heart and the corrected version
are shown with running mean and standard deviation which were computed using a
running average from 60 images (Figure \ref{fig:all_dice_06}). The dice scores
of all hearts, excluding the first (which had a dice score of 0.2) are shown in
figure \ref{fig:all_dice_06}. From the 300th heart onward, the vast majority
of hearts have a dice score above 0.98 (Figure \ref{fig:all_dice_09}) with a
few outliers having dice scores between 0.9 and 0.95 and just two extreme
outliers having dice scores below 0.7 (Figure \ref{fig:all_dice_06}).

\begin{figure*}
\centering
\begin{subfigure}{\textwidth}
  \centering
  \includegraphics[width=9cm]{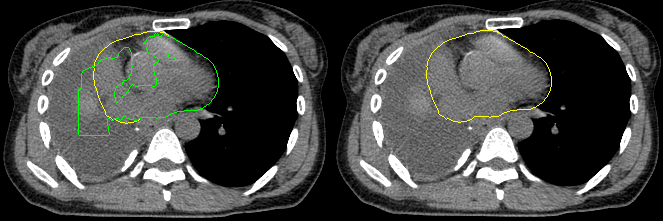}
    \caption{Outlier a: Dice score 0.67}
  \label{fig:outlier_a}
\end{subfigure}
\begin{subfigure}{\textwidth}
  \centering
  \includegraphics[width=9cm]{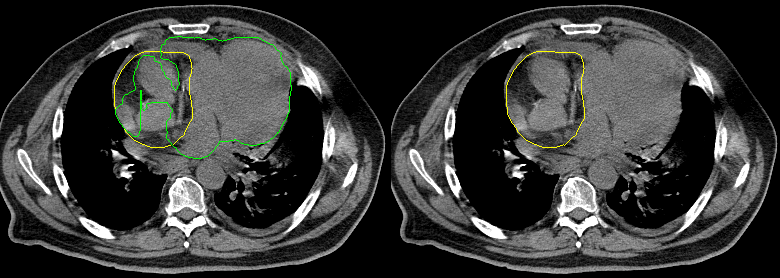}
  \caption{Outlier b: Dice score 0.68.}
  \label{fig:outlier_b}
\end{subfigure}
    \caption{Outliers a and b. For each heart two axial slices are shown with
contours overlaid. On the left the model prediction is shown in green with the
user corrected heart in yellow.  On the right only the corrected heart is
shown. For both the outliers with low dice, the tumour was located adjacent to
the heart.}
\label{fig:outliers}
\end{figure*}

Axial slices from the outliers a and b labelled in figure \ref{fig:all_dice_06}
are illustrated in figure \ref{fig:outlier_a} and figure \ref{fig:outlier_b}
respectively. These two hearts were the only ones out of the last 600 annotated
which had a dice score lower than 0.9. In both cases the model appeared to be
confused by a large tumour adjacent to the heart. 

To show the model progression more clearly, in figure \ref{fig:all_dice_09} the
y-axis minimum is raised to 0.9, which includes most of the values. Although
there are fluctuations, the mean dice score shows a trend of continuous
improvement as more images are correctively delineated (Figure
\ref{fig:all_dice_09}).

We found strong agreement between the hearts contoured in RootPainter3D and
Eclipse (Table \ref{table:last10dice}). All of the last 10 hearts in RootPainter3D
had higher agreement with the manual Eclipse delineations after the corrections
were assigned.

\begin{figure*}
\centering
\begin{subfigure}[t]{.464\textwidth}
  \centering
  \includegraphics[width=\textwidth]{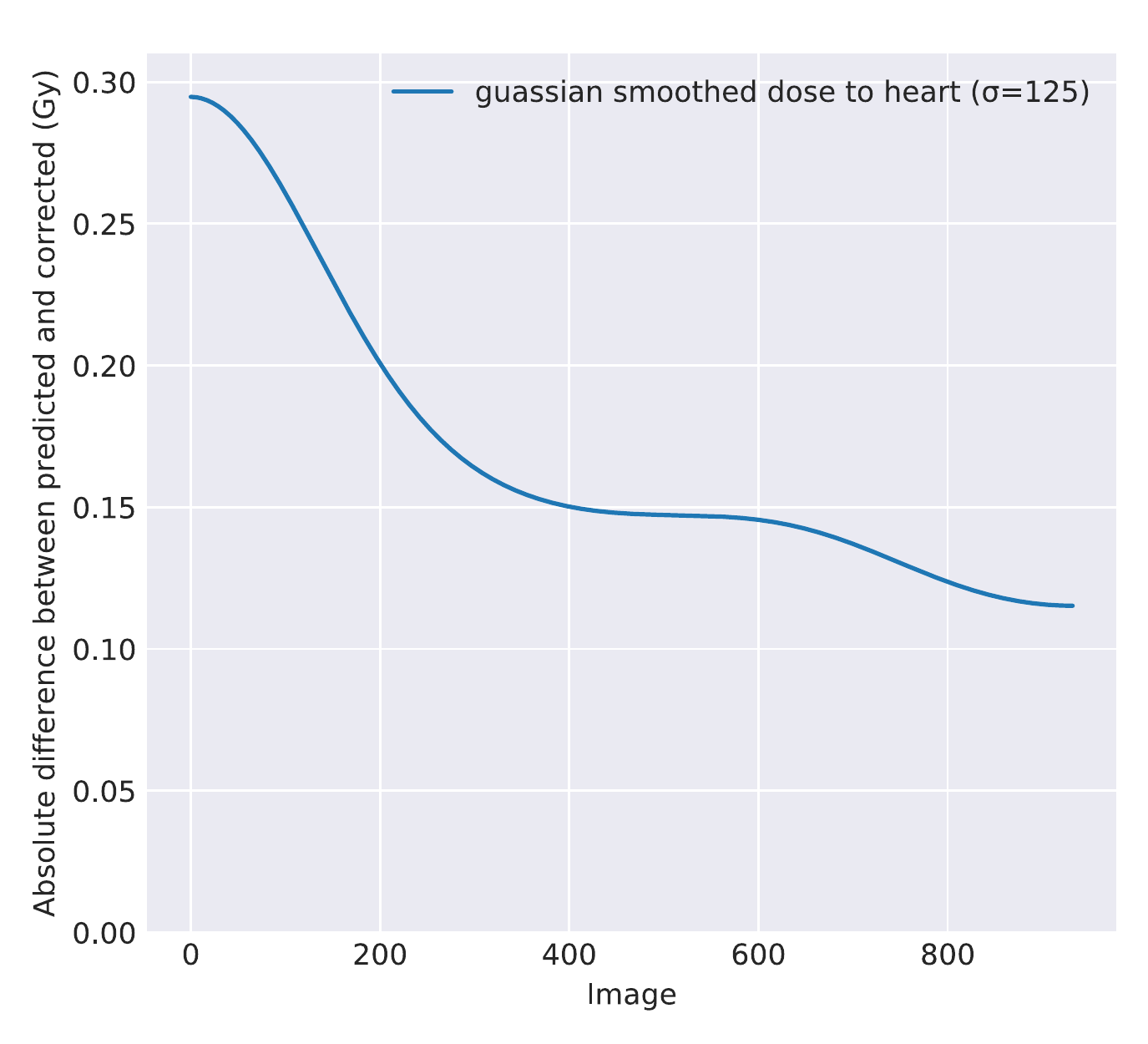}
    \caption{Running mean using a Gaussian kernel. The mean difference drops from ~0.3 to ~0.12 Gray as more images are correctively annotated.}
  \label{fig:dose_guassian_std_120}
\end{subfigure}
\hspace{0.1cm}
\begin{subfigure}[t]{.47\textwidth}
  \centering
  \includegraphics[width=\textwidth]{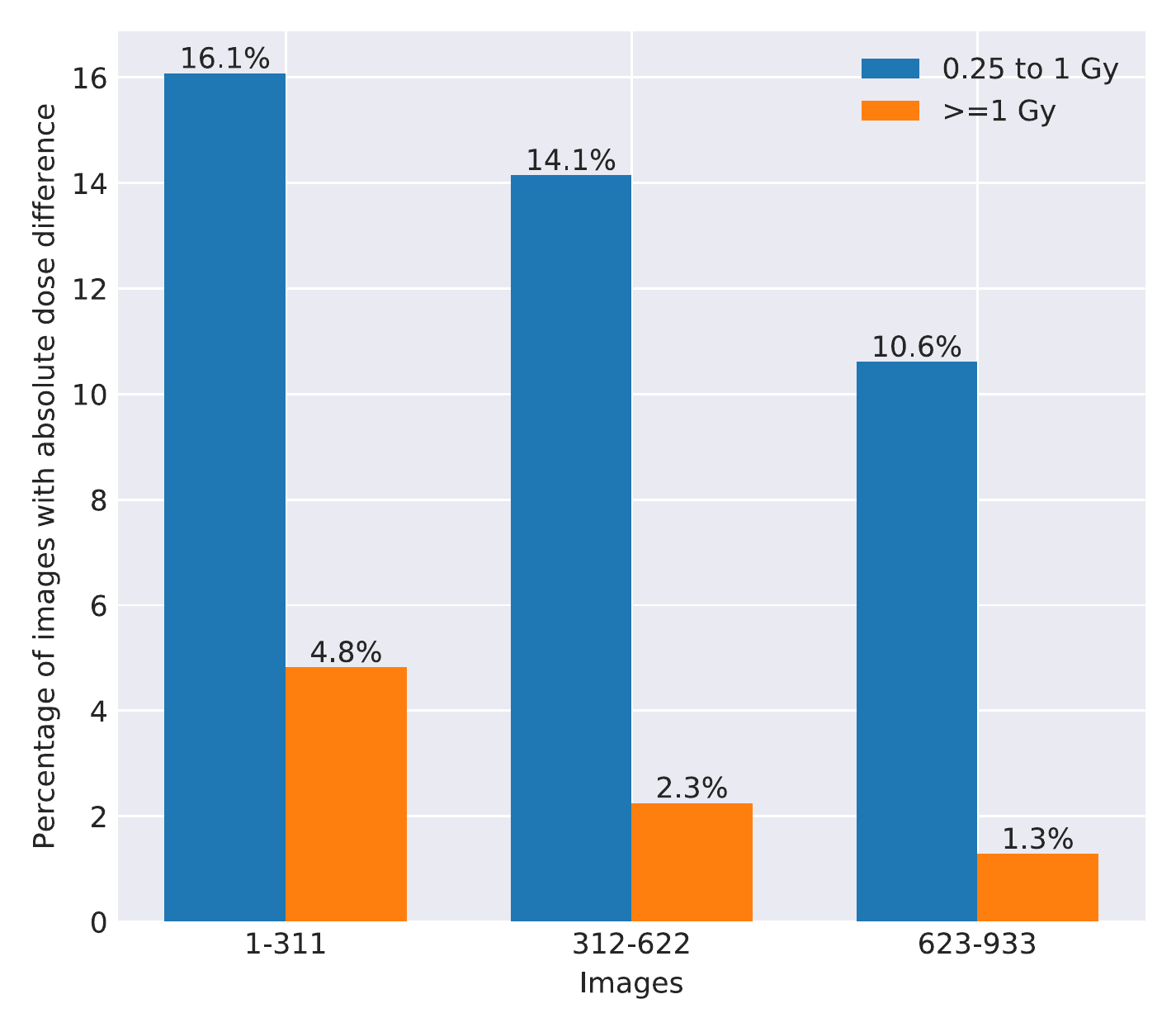}
  \caption{Percentage of hearts with absolute differences from 0.25 to 1 Gray and over 1 Gray for images 1-311, 312-622 and 623 to 923.}
  \label{fig:three_bars_dose_group}
\end{subfigure}
\caption{Absolute dose difference between predicted and corrected contours. The dose differences decrease as more images are annotated correctively.}
\label{fig:dose_difference}
\end{figure*}

Figure \ref{fig:dose_guassian_std_120} shows the error measured in mean dose for
automatically predicted heart structure drops as more images are correctively
delineated. As shown in \ref{fig:three_bars_dose_group} only 4 of the model
predicted hearts in the last 300 result in an error in dose above 1 Gray.

% 1 - dice with corrected heart vs eclipse 0.957
% 1 - dice with seg heart vs eclipse 0.945
% 2 - dice with corrected heart vs eclipse 0.959
% 2 - dice with seg heart vs eclipse 0.947
% 3 - dice with corrected heart vs eclipse 0.9522
% 3 - dice with seg heart vs eclipse 0.940
% 4 - dice with corrected heart vs eclipse 0.944
% 4 - dice with seg heart vs eclipse 0.930
% 5 - dice with corrected heart vs eclipse 0.952
% 5 - dice with seg heart vs eclipse 0.951
% 6 - dice with corrected heart vs eclipse 0.961
% 6-  dice with seg heart vs eclipse 0.954
% 7 - dice with corrected heart vs eclipse 0.952
% 7 - dice with seg heart vs eclipse 0.948
% 8 - dice with corrected heart vs eclipse 0.961
% 8 - dice with seg heart vs eclipse 0.957
% 9 - dice with corrected heart vs eclipse 0.938
% 9 - dice with seg heart vs eclipse 0.935
% 10 - dice with corrected heart vs eclipse 0.946
% 10 - dice with seg heart vs eclipse 0.941
% mean corrected_vs_eclipse_dice 0.952
% mean corrected_vs_predicted_dice 0.991
% mean predicted_vs_eclipse_dice 0.945

\begin{table}[H]
\caption{Dice scores between hearts delineated using RootPainter3D and Eclipse,
         showing strong agreement between methods. However, the RootPainter3D predicted and
         corrected contours are more similar to each other than either one is to the
         Eclipse contours. \textit{pred} refers to RootPainter3D predicted contours and \textit{cor} refers to the RootPainter3D corrected contours.}
\begin{tabular}{|l|l|l|l|}
    \hline
     & Eclipse vs pred & Eclipse vs cor & pred vs cor \\ \hline
mean & 0.945               & 0.952                & 0.991  \\ \hline
std  & 0.008                & 0.007                & 0.005  \\ \hline
\end{tabular}
\label{table:last10dice}
\end{table}


\subsection*{Discussion}
    \label{sec:discussion}

%Although we identified some disconnected components that the user forgot to
%correct, we computed the dice and found under no circumstances was the
%agreement in dice score less than 0.99 and the mean was above 0.9999,
%indicating the missed disconnected components were irrelevant.

The strong agreement with the manual contours shown in table
\ref{table:last10dice} supports our first hypothesis that RootPainter3D will
result in accurate contours. Prior heart auto-contouring studies have observed
a mean dice score of 0.925 between model predictions and manual delineation
\cite{feng_deep_2019}. \cite{yang_autosegmentation_2018} obtained a dice score
of 0.931 when measuring the disagreement between multiple annotators,
indicating the difference between RootPainter3D delineations and our manually
created delineations may be less than the difference between two annotators,
even before corrections are assigned.

The decrease in per-image contouring duration as more hearts are annotated
(figure \ref{fig:duration_with_eclipse}) and significantly faster contouring
for the last 10 hearts (Figure \ref{fig:two_plots_start_end} support our second
hypothesis that RootPainter3D will offer substantial time savings.

In Eclipse, only one in every three slices was manually delineated, with the
others created using interpolation. Even though in RootPainter3D all slices
were manually corrected, the contouring time was still more than three times
faster than Eclipse (Figure \ref{fig:two_plots_start_end}). This discrepancy between
the number of slices delineated vs interpolated also explains why RootPainter3D
was slower than eclipse during the initial period of annotation (Figure \ref{fig:two_plots_start_end}).

The dice scores between the RootPainter3D corrected contours and Eclipse
contours are lower than the dice scores between the RootPainter3D predicted
contours and corrected contours (Table \ref{table:last10dice}).

As the predicted and corrected contours are not done independently, it is
expected that they will have higher agreement than two independently created
contours. For quantifying errors in an existing delineation, there are some
cases where this may be a strength.  When delineating from scratch in noisy images
and areas with little contrast, a drop in dice may be caused by
differences in delineation along ambiguous border regions. With
corrective-annotation, the annotator focuses on clear errors and the measured
dice is therefore an indication of how often the system makes such clear errors
and these may be more relevant to quantify. For creating datasets for machine
learning, corrective-annotation could therefore reduce label noise consisting
of natural-perturbations. Although deep-learning is 
relatively robust to label noise \cite{rolnick_deep_2018}, perturbations along
boundaries are particularly problematic for U-Net model training
\cite{heller_imperfect_2018}.

In prior studies, correcting contours, as opposed to contouring from scratch has
been shown to increase consistency and reduce inter-observer variation
\cite{deeley_segmentation_2013}.

The two outliers with low dice score were in patients with anatomical
abnormalities that will also likely be correlated to high dose to the heart
(Figure \ref{fig:outliers}). This is a potential limitation of using fully-automatic
contouring for dose-response modelling and shows the importance of carefully
reviewing the results of such models in clinical applications.

A typical limitation of auto-contouring studies is the small number of scans
used for evaluation, with less than 100 being typical
\cite{schreier_generalization_2020}. By evaluating our contouring software on
over 900 hearts we were able to identify outliers that would have otherwise
been missed (Figure \ref{fig:outliers}). It is clear from figure
\ref{fig:all_dice_06} that smaller datasets may exclude such outliers by
chance, resulting in over-optimistic characterisations of performance on unseen
data.

%We speculate that corrective
%annotation allows a more precise measurement of the true delineation error, as
%opposed to ambiguous boundary region.

% Jens: Can we make a fairer comparison by say evaluating errors only in the
% corrected region? E.g. surface distance or something like that. This might
% allow us to say compare dice of rootpainter vs cynthia (eclipse and
% rootpainter) and cynthia vs cynthia. If rootpainter vs cynthia is lower on
% average, then it is better than her in repeating her own delineation.

% Need to discuss this in more details with Jens - Might be wrong.
% The consequences of labelling regions with errors for the training of fully
% automatic models are not yet entirely clear. \cite{czolbe_is_2021} found
% inconsistent results when using active learning to select training examples
% with high error for the training set.  Their approach provided similar benefits
% to random sampling which they attribute to a potential focus on more ambiguous
% regions and uncertainty in the data and \cite{smith_rootpainter_2020} did not
% find a significant difference between the models trained correctively and those
% using dense annotations.
% 

%%% Results in context to prior work and another discussion of IML

Despite editing auto-contours being identified as a barrier to the adoption of
automatic methods \cite{yang_autosegmentation_2018}, our results, in alignment
with previous studies \cite{lustberg_clinical_2018, vaassen_evaluation_2020,
kiser_novel_2020}, demonstrate that correcting auto-contours saves time
compared to a standard clinical workflow.  In prior work, time-savings have been
demonstrated, even when the predicted structures are of particularly low
quality, with a dice score as low as 0.46 \cite{gooding_comparative_2018}.

The benefits of corrective-annotation to routine clinical contouring are clear
as correcting inaccurate contours is already essential to optimise treatment
planning \cite{ezzell_guidance_2003, mackie_image_2003}. 
In addition to being usable by non-experts \cite{dudley_review_2018} and
providing performance improvements in comparison to fully-automated systems
\cite{holzinger_interactive_2019}, IML can provide trust and quality control
benefits \cite{michael_interactive_2020} which are of particular interest in a
radiotherapy context. 

The IML process (Figure \ref{fig:iml_diagram}) provides feedback to the
annotator on how much data is necessary to train a model to a given level of
accuracy and due to their extended exposure to model behaviour, provides
physicians with control of model training and insight into the
strengths and limitations of an AI system, attributes needed to ensure clinical
adoption \cite{cai_hello_2019}.

\cite{ramos_interactive_2020} characterise IML systems as being used for
task-completion or model-building. As opposed to \cite{smith_rootpainter_2020},
who evaluated IML with corrective-annotation for model-building, in this study
we also demonstrate the potential for assisting in task-completion.

%\subsection{Limtiations}

We used mean heart dose computation accuracy to measure model quality. Although
it brings our results closer to a value familiar to the working radiation
oncologist, the utility of mean heart dose for understanding cardiac toxicity
has been called into question by recent studies \cite{hoppe_meaningless_2020}.
With faster contouring capabilities, IML systems such as RootPainter3D provide
capabilities to contour more structures in less time, making it possible to
delineate organ sub-structures in a larger cohort of patients.

We motivated the omission of data-augmentation based on results that indicate
the benefits may be dataset specific \cite{isensee_nnu-net_nodate}.  More
recently, the same authors conducted a more thorough analysis, providing
stronger indications that a sensibly designed data-augmentation procedure would
provide benefits across datasets and organs \cite{isensee_nnu-net_2021}. Thus
we expect the addition of data-augmentation would further improve our results.

Similarly to \cite{brouwer_assessment_2020}, we observed that many time
consuming contour adjustments were minor changes along the boundary which are
in many cases unlikely to have a large impact on dose. We agree with
\cite{brouwer_assessment_2020} that there is a need for methods that guide
delineators in determining which corrections are meaningful for dose-planning.

% This is what \cite{brouwer_assessment_2020} says:
% """ Since many adjustments were done within 2 mm, the question arises to what
% degree contour adjustments are clinically meaningful. As other studies have
% shown, variability in contouring does not result in significant dosimetric
% differences for the majority of organs [4], [24]. A threshold for meaningful
% contour adjustments could possibly be set [22], and RTTs trained to accept
% contours in cases not exceeding this threshold. Practical tools to guide the
% user in this decision-making process will enable further automation of
% contouring practice. Further research could also consider the dosimetric
% impacts of the edits made within this cohort. """

The combination of electronic health records, stored dose matrices and CT scans
from record and verify systems has great potential to increase our
understanding of radiation dose effects for normal tissues
\cite{vogelius_harnessing_2020}. \cite{abravan_radiotherapy-related_2020}
investigated the relationship between radiation exposure and lymphopenia for a
cohort of 901 patients by registering all scans to a single reference patient.
Our proposed method would compliment such studies by enabling efficient
segmentation of structures of interest for each individual patient CT scan,
potentially mitigating errors that may result from the registration process.

Our software is made available open source and we emphasise that with a
restarted training process, the same application can learn to segment other
organs of interest in a research setting. As we started training from random
weights and use a fairly generic U-Net architecture, we expected the
performance to be similar.

In conclusion, our results demonstrate the benefits of continual-learning
with correction-annotation, by showing how contouring time can be continually
reduced whilst maintaining accuracy. It may be some time before such
systems can be made widely available to clinics, as medical devices utilising
continual-learning are yet to be approved by the FDA \cite{lee_clinical_2020,
cruz_rivera_guidelines_2020}. 

We also only used a single physician for our experiments and the contours were
collected for research purposes, rather than as part of a routine procedure.
The risks when working with new data and multiple clinicians will need to be
carefully considered before such systems can be adopted in the clinic, and new
tests may need to be designed to ensure robustness and quality
\cite{vokinger_continual_2021}.

\FloatBarrier % dont let the images go into the Acknowledgements.

% Discussion. 

% ### Discussion

% See yang_autosegmentation_2018 Figure: 2
% "Qualitative display of inter-rater differences in organ-at-risk segmentations."
% Shows how much varaiblity is due to inter-rater differences. We expect
% That a corrective approach will reduce that variation.
% But our data suggests the user still spent an inordinate amount of time
% fixing minor errors, similar to interoberver variation
% further work is required to show which errors matter

% Mention that open source allows users to customize the software
% to their specific hardware setup.

% ##### QA bit....
%## Clinical contouring optimisation: - Can we use the time savings to further improve QA???
% And how would using such a system increase the burdon on the clinic in terms of consistency 
% and technical expertise???

%quantitative analysis of real-time peer review quality assurance rounds incorporating direct physical examination for head and neck cancer radiation therapy. Int J Radiat Oncol Biol Phys 98:532-540, 2017  
%Marks LB, Adams RD, Pawlicki T, et al: Enhancing the role of case-ori- ented peer review to improve quality and safety in radiation oncology: Executive summary. Pract Radiat Oncol 3:149-156, 2013  
%Cox BW, Kapur A, Sharma A, et al: Prospective contouring rounds: A novel, high-impact tool for optimizing quality assurance. Pract Radiat Oncol 5:e431-e436, Sep. 2015 

\subsection*{Acknowledgements}

We thank Thomas Carlslund and Kurt Nielsen for IT infrastructure support and
Agata Wlaszczyk for proofreading. We would also like to thank Katrin Elisabet
Håkansson, Mirjana Josipovic and Emmanouil Terzidi for feedback on early
versions of the software and Deborah Anne Schut for
feedback on the heart contouring procedure. We also thank Mikkel Skaarup for
feedback on experimental design and gratefully acknowledge our financial
support from Varian Medical Systems and the Danish Cancer Society (grant no
R231-A13976).

\bibliographystyle{abbrv}
\bibliography{references.bib}

\end{document}